\documentclass{article}

\usepackage{microtype}
\usepackage{graphicx}
\usepackage{booktabs} 

\usepackage[justification=centering]{subfig}
\usepackage{ulem}
\renewcommand{\sout}[1]{\unskip}
\usepackage{xurl}
\usepackage{hyperref}

\usepackage[accepted]{icml2023}

\usepackage{amsmath}
\usepackage{amssymb}
\usepackage{mathtools}
\usepackage{amsthm}
\usepackage{natbib}

\usepackage[capitalize,noabbrev]{cleveref}

\theoremstyle{plain}

\theoremstyle{definition}

\theoremstyle{remark}

\usepackage[textsize=tiny]{todonotes}

\icmltitlerunning{CHILLI: A data context-aware perturbation method for XAI}

\begin{document}

\twocolumn[
\icmltitle{CHILLI: A data context-aware perturbation method for XAI}



\icmlsetsymbol{equal}{*}

\begin{icmlauthorlist}
\icmlauthor{Saif Anwar}{yyy}
\icmlauthor{Nathan Griffiths}{yyy}
\icmlauthor{Abhir Bhalerao}{yyy}
\icmlauthor{Thomas Popham}{yyy}
\icmlauthor{Mark Bell}{comp}

\end{icmlauthorlist}

\icmlaffiliation{yyy}{Department of Computer Science, University of Warwick, Coventry, United Kingdom}
\icmlaffiliation{comp}{TRL, Wokingham, United Kingdom}

\icmlcorrespondingauthor{Saif Anwar}{saif.anwar@warwick.ac.uk}

\icmlkeywords{Machine Learning, ICML}

\vskip 0.3in
]



\printAffiliationsAndNotice{} 

\begin{abstract}
  The trustworthiness of Machine Learning (ML) models can be difficult to assess, but is critical in high-risk or ethically sensitive applications. Many models are treated as a `black-box' where the reasoning or criteria for a final decision is opaque to the user. To address this, some existing Explainable AI (XAI) approaches approximate model behaviour using perturbed data. However, such methods have been criticised for ignoring feature dependencies, with explanations being based on potentially unrealistic data. We propose a novel framework, CHILLI, for incorporating data context into XAI by generating contextually aware perturbations, which are faithful to the training data of the base model being explained. This is shown to improve both the soundness and accuracy of the explanations.
  \end{abstract}

\section{Introduction}

Machine Learning (ML) and Artificial Intelligence (AI) are increasingly being used to tackle problems in a variety of domains because of their prodigious performance in automated decision-making. Some of these domains have high associated risks, such as financial systems \cite{alaraj_classifiers_2016, byanjankar_predicting_2015}, healthcare \cite{lodhi_predicting_2017, mikalsen_unsupervised_2018} and criminal justice \cite{rigano_using_2019}. Incorrect decisions in these scenarios can have significant repercussions, and making decisions with intentional or inadvertent biases can lead to discrimination and other social consequences \cite{reuters_amazon_2018, Sweeney_2013}. Therefore, it is essential that the decisions made by an ML model are trusted before being acted upon. The foundation of such trust is dependent on both developers and end users understanding the reasoning behind a model's decisions.

Due to the complexity of many ML techniques, they are often treated as a `black-box' where the reasoning for a prediction can be  difficult to ascertain. Such understanding would allow users to better detect biases in data, assess the vulnerabilities of a model, and ensure a model meets any regulatory standards \cite{goodman_european_2017} and societal expectations. Explainable AI (XAI) methods aim to increase confidence in AI systems, supporting their acceptance and wider adoption. While the use of XAI terminology varies, we define \emph{explainability} as providing evidence or reasoning for all outputs via an explanation, \emph{interpretability} is the notion that all explanations must be understandable to users, and \emph{faithfulness} is a measure of how accurately an explanation reflects the behaviour of an AI system. 


While some ML models are inherently interpretable \cite{sudjianto_designing_2021}, e.g., decision trees, where the behaviour of a model can implicitly be explained \cite{breiman_classification_2017}, other ML techniques require explanations to be generated separately.
Post-hoc XAI attempts to form explanations after a predictive model has been learnt. Such approaches are often model-agnostic and applicable to a range of ML techniques \cite{goldstein_peeking_2014, molnar2022, ribeiro_why_2016}. However, evaluations of model-agnostic approaches have shown that, just as ML models are adapted to their context, XAI systems should also be adapted to the appropriate deployment domain \cite{zhang_why_2019, sokol_blimey_2019}. 

It is often challenging to interpret context from numerical data representing quantitative information, features and signal values. For example, a value of $0.5$ may represent a probability of $50\%$ or a value in the range $[0,10]$. This is a common problem in XAI, where feature values are used to explain predictions without contextual knowledge of the data \cite{sokol_blimey_2019, zhang_why_2019}. Earlier works \cite{lieberman_out_2000, selker_context-aware_2000} discuss the importance of context sensitivity for computer systems, which is crucial for XAI in understanding complex ML models. An XAI framework requires underlying domain knowledge of numerical data, to incorporate the appropriate semantics into the explanation. In this paper, we explore the effects of incorporating contextual domain knowledge into XAI, highlighting its importance when explaining predictions. We demonstrate this by evaluating the interpretability and faithfulness of explanations in an intuitive and quantitative manner. The contributions of this paper are as follows.
\begin{itemize}
\item We present an analysis of an existing XAI framework, namely LIME \cite{ribeiro_why_2016}, to illustrate the impact on interpretability and faithfulness of explanations when  data context is disregarded.
\item We propose a method for incorporating data context into a post-hoc XAI framework when using a proxy model.
\item Finally, an algorithm is presented for generating local contextually aware data samples for use when fitting a proxy model.
\end{itemize}

\section{Related Work}
\begin{figure}[!t]
	\centering
	\includegraphics[width=\linewidth]{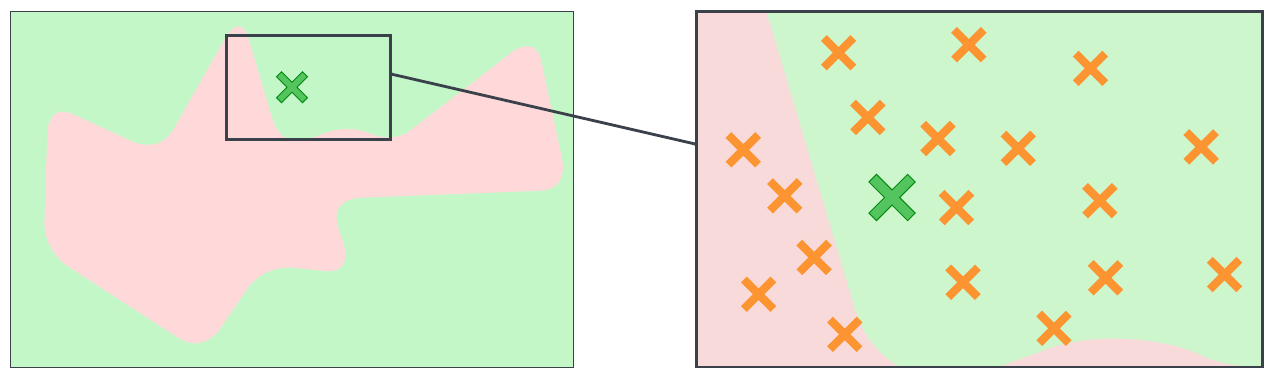}
  \hspace{0.5\linewidth}\subfloat[\label{global}]{}
  \hspace{0.5\linewidth}\subfloat[\label{local}]{}
	\caption{\label{fig:GlobalvLocal} (a) The global decision space, with red and green representing decision regions for two classes. (b) The local area around an instance, with a set of perturbations shown as orange crosses.}
\end{figure}

XAI methods either produce global or local explanations \cite{Mohseni_Zarei_Ragan_2021}. The former explain model behaviour overall \cite{Wang_Liu_Liu_Chen_Zhu_Guo, Lakkaraju_Bach_Leskovec_2016}, whereas the latter focus on a small area of the decision space, such as around a particular data instance \cite{ribeiro_why_2016, Ribeiro_Singh_Guestrin_2018, Zeiler_Fergus_2013, Baehrens_Schroeter_Harmeling_Kawanabe_Hansen_Mueller_2009}. This is illustrated in Figure \ref{fig:GlobalvLocal}, which shows the decision regions of a binary classifier, with the red and green areas representing different classes.
\subsection{Inherently Interpretable Models}

The structure of some ML models is inherently interpretable, e.g., linear regression where feature coefficients can be observed \cite{pml2Book}, and decision trees where the decision path can be traced. In these cases, an explanation is the base model itself, which is of course completely faithful to its own behaviour. As a result, such model types are often favoured in high-risk scenarios \cite{Rudin_2019}.

\subsection{Post-hoc Approaches}
It is not always be possible to use an inherently interpretable model, for example if a pre-built model requires explaining. Moreover, for some applications the best performing models are highly complex with large numbers of parameters \cite{simonyan_very_2015} and are consequently not inherently interpretable \cite{wickramanayake_towards_2021}. While there is increasing research into high performing inherently interpretable models \cite{sudjianto_designing_2021}, post-hoc XAI methods have been developed to explain existing models. These are typically model-agnostic, using only input and output data to understand model behaviour. Some methods use counterfactual explanations, by highlighting the consequence of modifying an input on a prediction \cite{verma_counterfactual_2020}, while others, such as Shapley values \cite{messalas_model-agnostic_2019} or proxy models, present contributions of features towards a prediction as an explanation. 

Proxy models approximate the behaviour of a base model in an interpretable form, such as a decision tree \cite{schmitz_ann-dt_1999} or linear regression model \cite{ribeiro_why_2016}. The proxy model is used as an explanation without sacrificing predictive performance of the base model. This is achieved by fitting a simpler proxy model to the base model predictions. A global proxy model would approximate the base model behaviour in all areas to increase interpretability, however this may not be sufficiently faithful because of oversimplification. It is generally accepted that there is a trade-off between faithfulness and interpretability \cite{dosilovic_explainable_2018}. Therefore, some XAI approaches use a local proxy model fit to the neighbourhood of an instance being explained. This reduces the coverage of the approximation, and so a more faithful explanation, yet with low complexity may be formed \cite{Wood-Doughty_Cachola_Dredze_2021}.

\subsection{Perturbation Based Methods}
\label{sec:perturbBasedMethods}

The Local Interpretable Model-Agnostic Explanations (LIME) method \cite{ribeiro_why_2016} explains the prediction for a given instance by fitting a proxy model in its locality. Since there may not be sufficient training data in a locality to fit a proxy model, algorithms such as LIME fit a proxy model to a set of synthetic perturbations of the instance being explained, as illustrated in Figure \ref{local}. 

In LIME, a set of perturbed inputs, $\mathcal{Z}$, is generated in the locality of an input instance, $x$, whose output prediction from some base model, $f$, is being explained. The base model is used to predict a set of target values, $f(\mathcal{Z})$, for the perturbations. A local proxy model, $g$, is then fit to this perturbed dataset. Non-categorical features are perturbed in LIME by sampling from a Normal distribution with mean and standard deviation estimated from the training data. Samples are taken from the center of the training data and then scaled around the instance \cite{Garreau_von}. Categorical features are perturbed by uniformly sampling from the distribution of feature values in the training data.

When fitting the proxy model according to some loss function, the loss contribution of each perturbation, $z \in \mathcal{Z}$, is weighted by a proximity measure, $\pi_x(z)$, according to some distance function, $D(x,z)$, to ensure the explanation is locally focused around $x$. By default in LIME, Euclidean distance is used and is calculated over all feature dimensions \cite{ribeiro_why_2016}. The proximity between two instances $\mathbf{p}$ and $\mathbf{q}$, is calculated as shown in Equation \ref{eq:proximity}, where $\sigma$ is a hyperparameter defining the locality of the explanation. 

\begin{equation}
    \label{eq:proximity}
    \pi_p(q) = exp{\left(\displaystyle\frac{-D^2(\mathbf{p}, \mathbf{q})}{\sigma^2}\right)}
\end{equation}


In a review of model-agnostic XAI approaches, \citet{molnar_general_2021} observe that perturbation-based methods tend to ignore feature dependencies by extrapolating in areas that are not representative of the original data distribution, and are therefore unknown to the base model \cite{molnar_model-agnostic_2021}. They also suggest that ignoring contextual constraints may lead to unrealistic data. For example, when perturbing a feature representing a person's age, the perturbation method must consider that values cannot be negative or unreasonably large \cite{molnar_general_2021}. An explanation  fit to such data will therefore not be faithful to the true behaviour of the base model, and so additional feature dependency information should be included \cite{molnar_model-agnostic_2021}. 


In this paper, we address the importance of feature dependence and propose a new framework, CHILLI, that incorporates dependency information using prior domain knowledge, and explore the effect on the faithfulness of explanations.

\subsection{Evaluating XAI Methods}
\label{sec:EvaluatingXAI}
A satisfactory explanation provides transparency, allowing users to understand decisions and how data was used. Quantifying properties such as transparency and interpretability is difficult since the desiderata of XAI include subjective properties relating to trust, ethics and understanding.

In this paper, we quantify the performance of proxy-model XAI approaches through the faithfulness of explanations. The faithfulness of a proxy model, $g$, used to explain a base model, $f$, can be calculated using an error metric to compare the predictions made by $f$ and $g$ on a set of input instances. For a global proxy model, the inputs used for evaluation may be from the training data used for the base model, $f$ \cite{Wood-Doughty_Cachola_Dredze_2021}, whereas for a local proxy model this may be a set of perturbations, $\mathcal{Z}$, around an instance of interest \cite{ribeiro_why_2016}. From a set of possible proxy models, $G$, the proxy model with the lowest error, $g$, is selected as the explanation since it is most faithful.

\section{Contextually Enhanced Interpretable Local Explainable AI}
In this section, we propose Contextually Enhanced Intepretable Local Explainable AI (CHILLI), an XAI framework that combines the contextually aware proximity measures and domain representative perturbation generation method presented below
to explain base model behaviour using local proxy models. CHILLI aims to satisfy potential contextual constraints and consider limitations of numerical data. Explanations are fit to perturbed data that is representative of the base model training data and is local to the instance being explained. CHILLI is based on LIME \cite{ribeiro_why_2016}, with modifications to the proximity calculations and perturbation generation methods.

\subsection{Contextually Aware Proximity Measures}
\label{sec:NewProximityMeasures}
Proximity in LIME is based on Euclidean distance (see above, Section \ref{sec:perturbBasedMethods}), irrespective of the feature type. However, if for some features the absolute difference between two values does not reflect their distance, this proximity measure becomes invalid. This is the case if, for example, the units are not equidistant, or a feature is not measured linearly, such as magnitude recorded on a logarithmic scale. Such distance measures are also unsuitable for cyclic or temporal features e.g., hour of day where raw values for 23:00 and 00:00 appear to be far apart, but domain knowledge informs us that they are consecutive.

We propose that the context of features should be considered independently by incorporating the scale and bounds of each feature to ensure the calculated distance is truly representative. Consider the points $\mathbf{p}$ and $\mathbf{q}$, represented by feature vectors of $d$ dimensions. Instead of using a generic distance function, such as Euclidean distance, the distance between the points, $D(\mathbf{p},\mathbf{q})$, is calculated individually for each feature dimension, $i$, using a specified distance function, $D_i$. The distance in each feature dimension is normalised, to allow for equal contribution, and averaged across all dimensions to give a single distance value. From this, a proximity measure can be calculated, as shown in Equation \ref{eq:newProximity}.
\begin{equation}
    \label{eq:newProximity}
    \pi_\mathbf{p}(\mathbf{q}) = exp{\left(\displaystyle\frac{-(\frac{1}{d}\sum_{i\in d}D_i(p_{i}, q_i))^2}{\sigma^2}\right)}
\end{equation}
Since the proximity measure is used when quantifying the performance of each explanation, $g$, an accurate proximity measure is essential to ensure the most faithful explanation model, $g$, is selected from the set of possible explanations, $G$.

\begin{figure}[!h]
\centering
  \includegraphics[width=\linewidth]{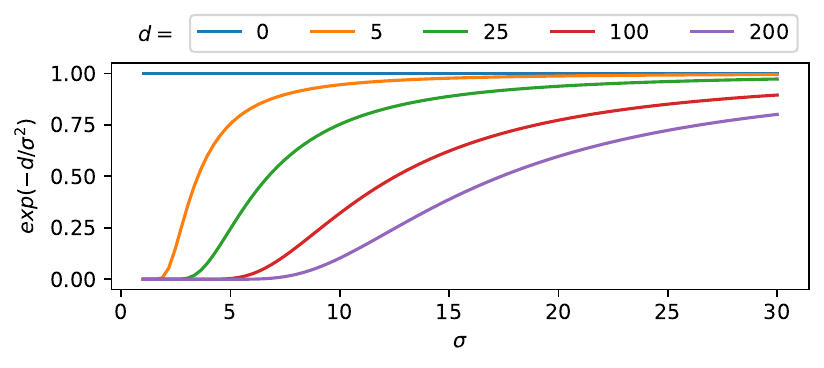}
  \caption{\label{fig:sigmaVariation} A comparison of the effect on proximity between two points of fixed distance, $d$, as the locality parameter, $\sigma$, is varied.}
\end{figure}

The value of the locality hyperparameter, $\sigma$, may be adjusted to vary the locality of an explanation in the model space. As $\sigma$ increases, the proximity tends to 1, as shown in Figure \ref{fig:sigmaVariation} for a range of distance values. All perturbations for a high enough value of $\sigma$, regardless of distance, will be assigned a proximity of 1 and are considered equally when selecting the best fit proxy model. Conversely, a smaller value of $\sigma$ will result in greater variation between proximities for perturbations of differing distance to the instance being explained. This leads to the selection of a proxy model that performs better on perturbations of closer proximity, thus reducing the locality of the explanation.

\subsection{Domain Representative Perturbation Generation}
\label{sec:NewPerturb}

Existing perturbation-based XAI methods, such as LIME, do not consider contextual knowledge when generating perturbations, which are the foundation for creating an explanation \cite{ribeiro_model-agnostic_2016, ribeiro_why_2016, zhang_why_2019, sokol_blimey_2019}. Perturbations in LIME are sampled uniformly from the entire feature space, and a proxy model fit to these will focus on all relationships in the feature space. Such an explanation is not local to the instance being explained, but is generalised to the overall model behaviour. Presenting this as a local explanation may be misleading regarding the behaviour of the model.

Such perturbations also ignore any bounds on feature values, such as `Age' which can only take positive values. Without considering these bounds, perturbations  may contain unrealistic values. Moreover, since features are perturbed independently, feature dependencies are ignored. For example, assume that as the population of a city grows, traffic congestion increases. Although these features are correlated,  ignoring feature dependancies may result in a perturbation combining lowered congestion with increased population. The omission of such dependencies may result in an unrealistic set of perturbations, leading to an explanation that does not describe the behaviour of the base model \cite{laugel_dangers_2019}. An XAI framework should generate perturbations that are representative of real-world data, such as the training data, and are local to the instance being explained.

We present a framework, CHILLI, for generating local and contextually conforming perturbations. Our method takes inspiration from the SMOTE algorithm \cite{chawla_smote_2002}. SMOTE is a well-regarded sampling technique used to generate synthetic data when there is class imbalance \cite{chawla_smote_2002}. A random point is selected from the minority class and its $k$-nearest-neighbours are located, for a predetermined number, $k$. One of these neighbours is selected uniformly at random and a synthetic datapoint is generated by linearly interpolating between the two points and uniformly selecting a random point on the line joining the two. This is repeated for different instances from the minority class until a specified degree of over-sampling has been achieved.

We use this approach to generate perturbations, and to ensure perturbations fall within realistic bounds, they are produced by interpolating between an instance being explained, $x$, and some other randomly selected instance, $x'$, in the training data. Each feature value is interpolated independently. For categorical features, the interpolated value is rounded to the nearest feature value.

To maintain the locality of the perturbations, the selection of $x'$ is from a probability distribution calculated using proximity (Equation \ref{eq:newProximity}) which is normalised for all data points such that $P(x'= x_i) = \pi_{x}(x_i)$ and $\sum_{x_i \in \mathbf{X}}P(x' = x_i) = 1$. As a result, perturbations are more likely to contain values in closer proximity to $x$. The process for generating a set of $N$ perturbations is outlined in Algorithm \ref{alg:PerturbationGeneration}.
\begin{algorithm}[!ht]
  \caption{\label{alg:PerturbationGeneration} Contextual Perturbation Generation}
  \begin{algorithmic}[1]
  \renewcommand{\algorithmicrequire}{\textbf{Input:}}
 \renewcommand{\algorithmicensure}{\textbf{Output:}}
	\REQUIRE Number of perturbations to generate, $N$; Data instance to perturb, $x$; Training dataset, $\mathbf{X}$
 	\ENSURE Set of perturbations $\mathcal{Z}$
    \STATE $F$ = Features of $x$

 	\STATE Initialise empty set of perturbations, $\mathcal{Z} = [$ $]$
 	\STATE Calculate $\pi_x(x^i)$ for each $x^i \in \mathbf{X}$
 	\STATE Assign a probability to each $x^i$ where $P(x'=x^i) = \frac{\pi_x(x^i)}{max(\pi_x(x^i) \forall x^i \in \mathbf{X})}$

	\WHILE{$ \vert \mathcal{Z} \vert \leq N$}
      \STATE Uniformly select some value between 0 and 1 $\rightarrow I$
	  \STATE Select some $x^i \in \mathbf{X}$ based on probability for each $x^i \rightarrow x'$ 
	    \FOR{$f$ in $F$}
    		\STATE $z_f$ = $x_f + I(x'_f - x_f)$
	    \ENDFOR
      \STATE $\mathcal{Z} = \mathcal{Z} \cup \{z\}$
	  \ENDWHILE
  \end{algorithmic}
\end{algorithm}



\section{Experimental Setup}
We compare the functionality and performance of our proposed method, CHILLI, with that of LIME \cite{ribeiro_why_2016}, to explore the effect of incorporating contextual information into XAI frameworks.

\subsection{Datasets}
\label{sec:Datasets}
Our evaluation uses the WebTRIS and MIDAS datasets. WebTRIS \cite{highways_webtris_2017}, recorded by Highways England, contains traffic data at 15 minute intervals for many motorway sites around England. We restrict the dataset to a single site (M6/7570A) between 01/01/2016 and 01/01/2017. A Support Vector Regressor (SVR) is trained on a subset of the available features describing date, time, average speed and number of vehicles of various sizes, to predict the `Total Volume' of traffic per time interval. The data distribution  in the individual feature dimensions is shown in Figure \ref{fig:webtrisPredictions}.

MIDAS \cite{met_office_midas_2022}, published by the UK Meteorological Office, records hourly weather observations at multiple locations across the UK. A Recurrent Neural Network (RNN) is trained on data containing various weather features from a station located at Keswick and 3 neighbouring stations (St. Bees Head, Shap and Warcop Firing Range) to predict `Air Temperature' at Keswick at a given time. It is expected that observations from surrounding areas will be related to the upcoming weather at Keswick, and therefore the data used from neighbouring stations is offset by 1 hour. The distribution of the training data is shown in Figure \ref{fig:midasPredictions}.

We can hypothesise about expected feature importance due to the linearity of the feature relationships against the target variable. Since LIME scales perturbations according to the covariance of the feature against the target variable, we explore the effect on explanation performance of removing generally linear features from the MIDAS data (namely, those describing relative humidity and dewpoint). 

\subsection{Forming Explanations}
Explanations containing a set of linear coefficients are produced using CHILLI and LIME for predictions made by a base model. The magnitude of each coefficient indicates the contribution of the corresponding feature towards the base model prediction, whilst the sign indicates the direction of the correlation between the feature and the target variable.

\begin{figure*}[!t]
  \centering
  \includegraphics[width=0.8\linewidth]{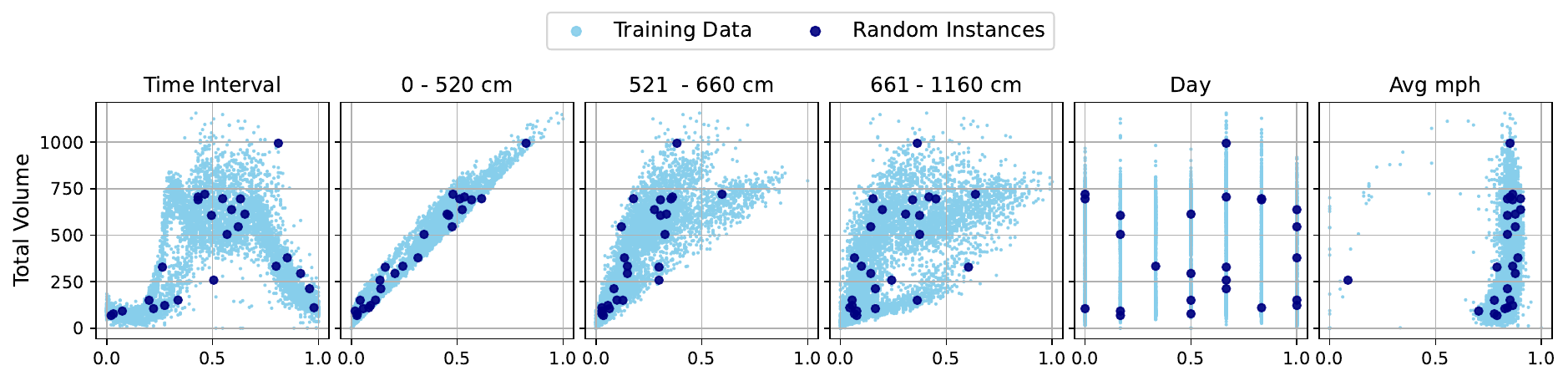}
  \caption{\label{fig:webtrisPredictions} Normalised WebTRIS training data shown in each feature dimension against the target `Total Volume' feature. The 25 instances selected uniformly at random for evaluation are shown as the dark blue points.}
\end{figure*}

\begin{figure*}[!t]
  \centering
  \includegraphics[width=\linewidth]{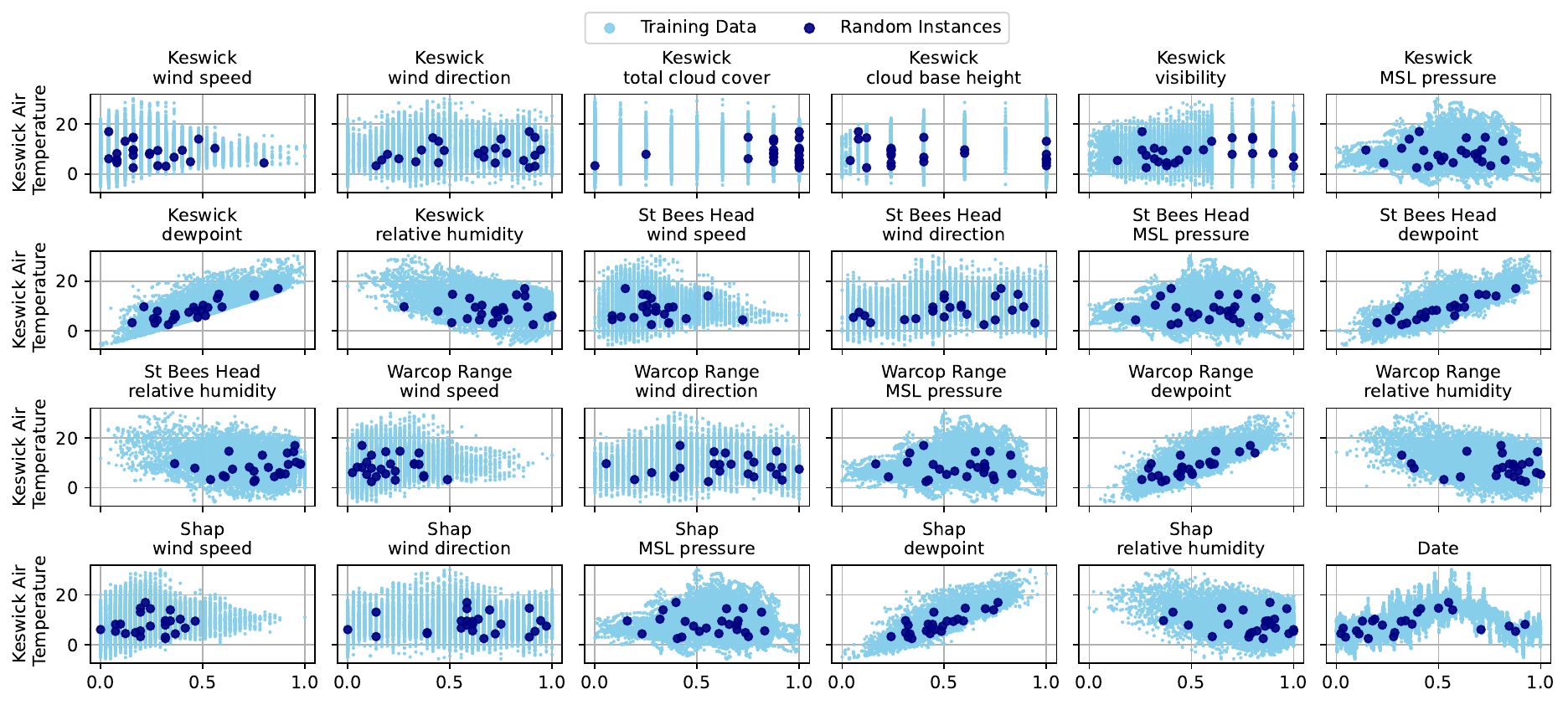}
  \caption{\label{fig:midasPredictions} Normalised MIDAS training data shown in each feature dimension against the target `Keswick Air Temperature' feature. The 25 instances selected uniformly at random for evaluation are shown as the dark blue points.}
\end{figure*}

We quantify the performance of an explanation using its error, which represents its faithfulness towards the base model. Explanations for WebTRIS predictions are quantified using RMSE and MAE is used for MIDAS predictions. We compare the error for explanations produced using CHILLI and LIME over 25 instances, selected uniformly at random.
The selected instances are shown in Figures \ref{fig:webtrisPredictions} and \ref{fig:midasPredictions}.

\label{sec:Results}

\begin{figure*}[!t]
  \centering
  \begin{tabular}{cc}
    \includegraphics[width=.75\linewidth]{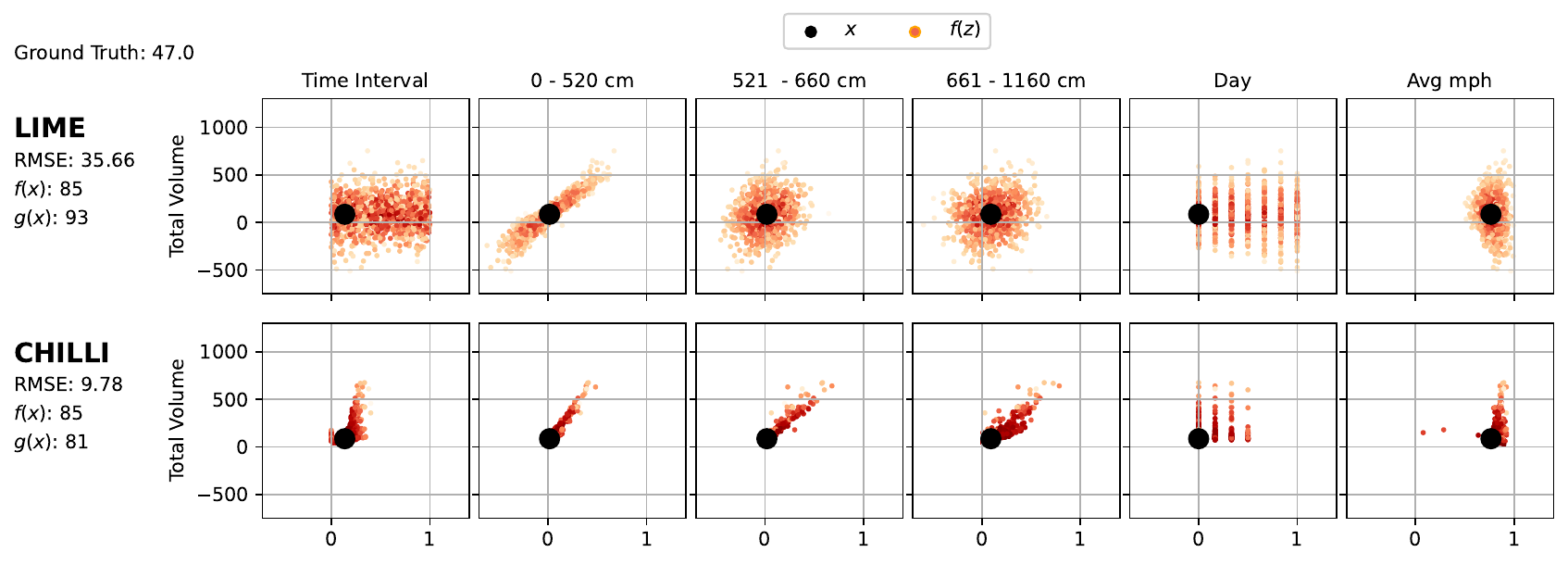} & \includegraphics[width=.24\linewidth]{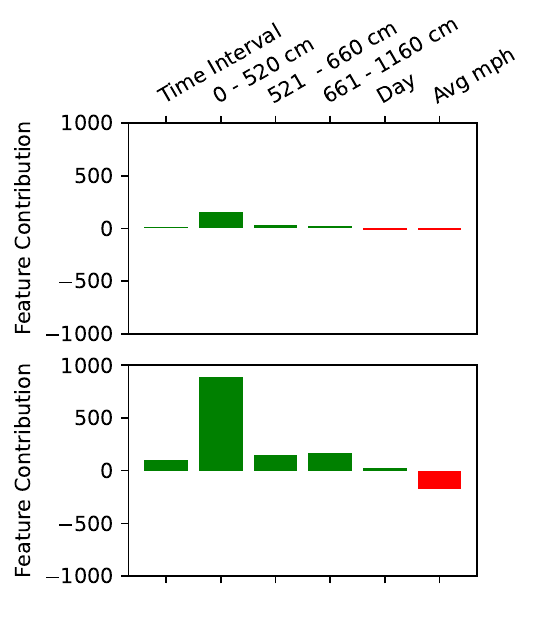}\\[\abovecaptionskip]
    \small (a) & \small (b) \\
  \end{tabular}
    \caption{(a) Perturbations (orange points) generated using LIME and CHILLI for a WebTRIS data instance, $x$. The predicted `Total Volume' for each perturbation from the base model, $f(z)$, is shown on the vertical axis. Opacity of perturbations represent proximity to $x$. (b) Explanations produced by CHILLI and LIME, showing the feature coefficients of the linear proxy model fit to the perturbations in (a) representing the contribution of each feature towards the predicted target value, $f(x)$.} \label{fig:webtrisResults}
  \end{figure*}

\section{Results \& Discussion}

In this section, we use LIME and CHILLI to fit a local proxy model which is used to explain a prediction produced by a base model for a given instance.

\subsection{Perturbation Generation}

Figure \ref{fig:webtrisResults} shows explanations produced by CHILLI and LIME alongisde the perturbations used to fit them for a prediction made by the SVR base model, $f$, for a randomly selected instance, $x$, from the WebTRIS test dataset. The instance is shown as the black point in each of its feature dimensions against the true `Total Volume' value, which is 47. The perturbations of the instance with the prediction of the target feature `Total Volume' from the base model, $f(z)$, are shown as orange points.

From a visual inspection of Figure \ref{fig:webtrisResults}a, it can be seen that the perturbations of features generated by LIME do not follow the data distribution shown in Figure \ref{fig:webtrisPredictions}. Moreover, since each feature is perturbed independently, feature values in a single perturbation do not consider feature dependencies, which leads to unrealistic perturbations being generated. For example, a perturbation may have a `Time Interval' of 03:00, but the value of `0-520cm' may correspond to the number of vehicles that would be observed at rush hour.

The bounds of features have also not been considered, 
as can be seen in Figure \ref{fig:webtrisResults}a where all non-categorical features exhibit perturbed values which fall outside the normalised range of $[0,1]$. Negative values of `0-520cm', `521-660cm' and `661-1160cm' imply a negative number of vehicles of the respective sizes passing in the corresponding time interval, which is not possible.
This leads to a set of perturbations that do not represent real-world data, and therefore do not represent the training data. The negative impact of such inappropriate perturbations can be observed from the predicted values from the base model, which often predicts a negative volume of traffic flow, which is also not possible.
An explanation that is fit on such perturbations will not correctly represent the true behaviour of the base model.

The opacity of each perturbation, shown in Figure \ref{fig:webtrisResults}a, signifies its calculated proximity weighting, $\pi_{x}(z)$, to the instance being explained. Perturbations which are further from the instance are sometimes assigned a higher  weighting than those which are closer. Since this indicates the contribution of each perturbation to the selection of the best fit linear proxy model, the produced explanation will not be locally focused around the instance being explained, and is instead a generalised explanation across all the perturbations.

On the other hand, the perturbations generated by CHILLI not only conform to the distribution of the training data shown in Figure \ref{fig:webtrisPredictions}, but are also realistic combinations of feature values that fall within the appropriate feature bounds. CHILLI also generates perturbations with greater density around the instance being explained, which can be seen in Figure \ref{fig:webtrisResults}a from the concentration of orange points around the instance. This is also the case for features of a cyclic nature, such as `Time Interval' in WebTRIS, where $0$ and $1$ are adjacent values. This leads to an explanation that is fit with greater emphasis on perturbations of closer locality. 

\subsection{Feature Contributions}

The linear proxy model with the lowest error when fit to the set of perturbations, $\mathcal{Z}$, and base model predictions, $f(\mathcal{Z})$, is selected as the explanation for the instance being explained. The explanations shown in Figure \ref{fig:webtrisResults}b indicate that CHILLI produces explanations with greater disparity between feature coefficients. It is expected for explanations produced by CHILLI to have larger feature coefficients than LIME, since the perturbations generated by LIME are based on a Normal distribution which naturally does not exhibit any linear correlation. Due to the covariance scaling of  LIME perturbations towards the training data, some features, such as `0-520cm' in the WebTRIS data, exhibit a strong linear correlation with `Total Volume' as shown in Figure \ref{fig:webtrisPredictions}. LIME recognises this and identifies it as the most significant feature contribution in its explanation. Similarly, features in the MIDAS data containing `dewpoint' and `relative humidity' have a linear correlation  with `Keswick Air Temperature' across their range of values, as can be seen from Figure \ref{fig:midasPredictions}. 

\begin{figure}[!t]
\centering  
  \includegraphics[width=\linewidth]{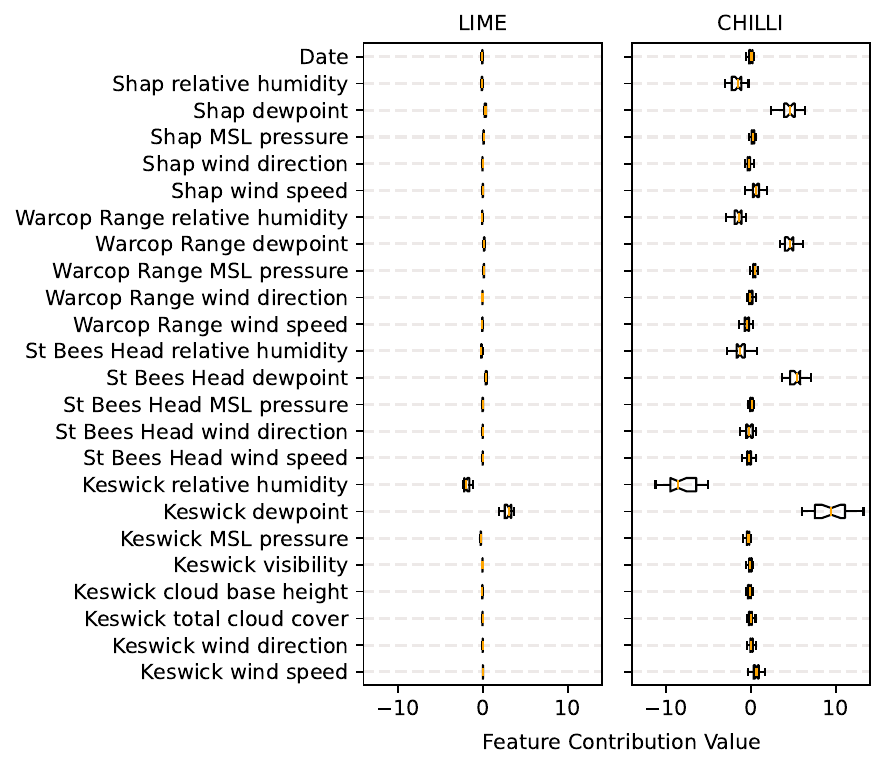}
  \caption{\label{fig:expVarianceLinear} Variation in feature contributions presented in explanations produced by LIME and CHILLI across the 25 instances shown in Figure \ref{fig:midasPredictions}. The median value and quartile ranges are shown for each feature.}
\end{figure}

Figure \ref{fig:expVarianceLinear} shows the variation in feature contributions in explanations produced for the 25 instances shown in Figure \ref{fig:midasPredictions}. Again, only generally linear features have noticeable contribution in the explanations produced by LIME. This is unsuitable since general feature trends are not relevant in a locally focused explaination. CHILLI produced explanations that are fit to perturbations local to the instance being explained, and recognises general linear trends in cases where the linear relationship is also present locally. However, CHILLI also highlights contributions from other locally impactful features, although they are not as significant as the generally linear features.

\begin{figure}[!t]
  \centering
  \includegraphics[width=\linewidth]{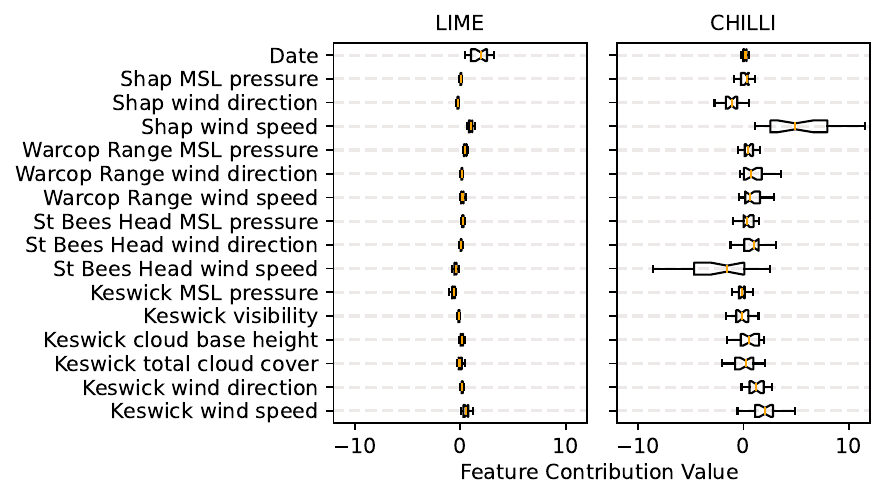}
  \caption{\label{fig:expVariance} Variation in feature contributions presented in explanations produced by LIME and CHILLI for the 25 instances shown in Figure \ref{fig:midasPredictions} when generally linear features are excluded. The median value and quartile ranges are shown for each feature.}
\end{figure}



Upon removal of generally linear features, there is greater variation in the explanations produced by CHILLI, as shown in Figure \ref{fig:expVariance}. Since LIME cannot detect local behaviour,
it performs poorly when locality is important, and does not identify any significant feature contributions due to the absence of general trends. CHILLI, on the other hand, is able to detect local trends and achieves significantly lower MAE, indicating that the explanations produced by CHILLI are more faithful to the true behaviour of the base model. 

\subsection{Explanation Faithfulness}

As noted in Section \ref{sec:EvaluatingXAI}, a lower error indicates a more faithful explanation. The explanation produced by CHILLI achieved a signifcantly lower RMSE than LIME on the perturbations shown in Figure \ref{fig:webtrisResults}. The explanation produced by CHILLI predicted the `Total Volume' of traffic for the instance to be 81 whilst LIME's explanation predicted 93. The lower error of the CHILLI explanation, combined with a prediction closer to the base model, supports the conclusion that CHILLI produces a more faithful explanation, and is more representative of the base model's true behaviour.

\begin{figure}[!t]
  \centering
    \includegraphics[width=\linewidth]{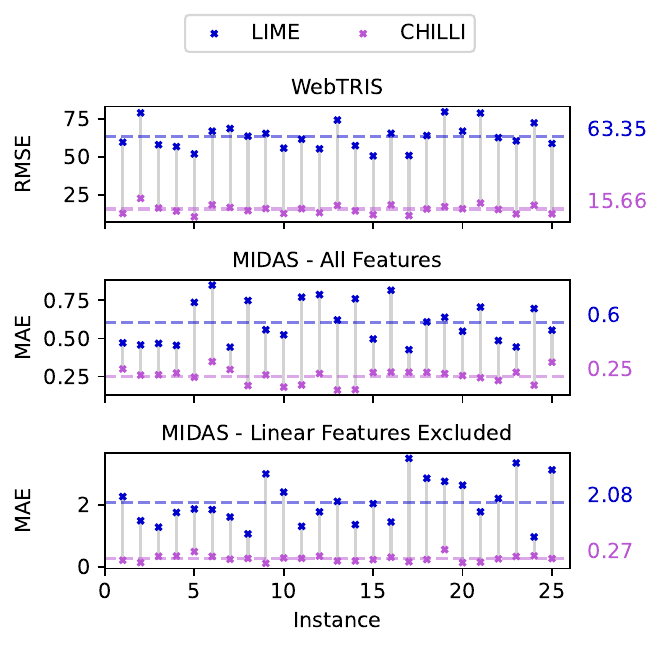}
    \caption{\label{fig:rmsecomparisonWebTRIS} Individual (crosses) and average (dashed line) error achieved by explanations from LIME and CHILLI for 25 randomly selected instances from WebTRIS and MIDAS.}
  \end{figure}

Figure \ref{fig:rmsecomparisonWebTRIS} shows a comparison of the error achieved by explanations produced by LIME and CHILLI for both  WebTRIS and MIDAS for the 25 instances shown in Figures \ref{fig:webtrisPredictions} and \ref{fig:midasPredictions}. Explanations were also produced for the MIDAS instances after removing generally linear features. In all explained instances, CHLLI achieves a lower error than LIME. The average error for each technique across all instances is also indicated in Figure \ref{fig:rmsecomparisonWebTRIS}. CHILLI leads to an average reduction in error of 75\% and 58\% on  WebTRIS and MIDAS respectively, with all features included. After removing generally linear features, the error of explanations produced by LIME increases significantly, whereas CHILLI maintains a similar  error since it captures other local trends. This leads to an average reduction in error of 87\% for CHILLI compared to LIME.
  

\subsection{Locality Hyperparameter Exploration}

\begin{figure}[!t]
  \centering
    \includegraphics[width=\linewidth]{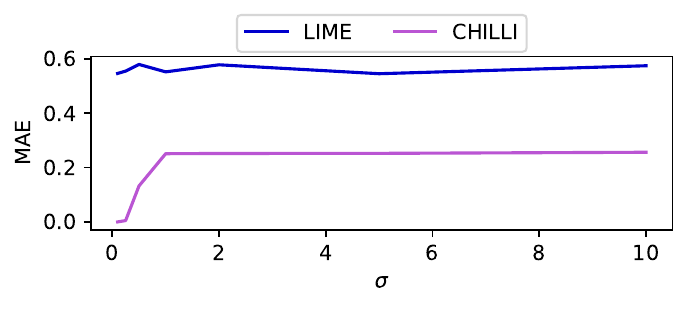}
    \caption{\label{fig:sigmaComparison} A comparison of the MAE achieved by explanations produced using LIME and CHILLI for a single instance of the MIDAS data, whilst varying the locality parameter, $\sigma$.}
  \end{figure}

The importance of accurate proximity measurements  can be understood by observing the effect of varying $\sigma$ on MAE. Figure \ref{fig:sigmaComparison} shows a comparison of the MAE achieved by LIME and CHILLI for a uniformly randomly selected instance from MIDAS, when using different values of $\sigma$. The MAE achieved by LIME is similar across all values of $\sigma$. Since LIME forms explanations that do not consider the local data context of the instance being explained, it is not expected for them to vary based on locality size. CHILLI achieves lower MAE for lower values of $\sigma$ before stabilising at higher values. As the defined locality increases, perturbations are considered that are further from the instance being explained. Features which do not exhibit linear relationships on a broader scale are difficult to describe using a linear proxy model. This leads to a worse performing explanation, since it attempts to generalise behaviour rather than explaining local trends, as with LIME. While MAE increases when using CHILLI, it still outperforms LIME since the intuition regarding perturbation generation is sound.

\section{Conclusion and Future Work}

In this paper, we explored the effect of incorporating contextual domain knowledge into a model-agnostic local perturbation-based XAI approach, namely LIME.
We proposed a method for contextually aware proximity measures, to ensure locality is accurately defined and constrained. We also proposed a  method for generating perturbations that consider the contextual limitations and dependencies of data features. These methods are combined into a new framework, CHILLI, for generating local explanations for black-box ML models, and we compared the functionality and performance of CHILLI with LIME.

Using the WebTRIS and MIDAS datasets, we demonstrated that LIME does not appropriately measure proximity between instances, resulting in an explanation which is not local to the instance being explained. Explanations generated by LIME were found to be generalised and only consider features with general linear trends. It was also found that LIME does not generate perturbations that are representative of the training data, and the perturbations contained unrealistic values.

CHILLI was shown to generate perturbations that are representative of the base model training data and are  local to the instance being explained. Therefore, CHILLI's explanations had relatively larger feature contributions compared to those produced by LIME. CHILLI consistently achieved a lower error, and therefore produced a more faithful explanation, across all explained instances compared to LIME.

Through empirical and intuitive evaluation of LIME and CHILLI, we conclude that incorporating contextual domain knowledge regarding data features used for generating explanations improves faithfulness, which may ultimately increase trust in both the explanation and explanation framework. In future work we will investigate how improving the performance of local explanations affects the overall trust in a model. We would also like to explore the efficacy of CHILLI when proxy models of a different form are used, such as decision trees or small-order polynomial regressors. 
\section*{Acknowledgements}

We gratefully acknowledge the funding provided by the UK Engineering and Physical Sciences Research Council (grant ref. EP/T517641/1) and TRL Ltd to support this iCASE with project code B.CSAA.0001.


\bibliographystyle{icml2023}
\bibliography{refs}



\end{document}